\documentclass{article}
\usepackage{natbib}
\usepackage{amsmath,graphicx}
\usepackage{booktabs}
\usepackage{soul, color}
\usepackage{float}
\usepackage[utf8]{inputenc}
\usepackage{authblk}
\usepackage{placeins}
\usepackage[top=1in, bottom=1in, left=1in, right=1in]{geometry} 
\usepackage{makeidx}  
\usepackage[utf8]{inputenc}
\usepackage{graphicx}
\usepackage[misc]{ifsym}
\usepackage{comment}
\usepackage{multirow}
\usepackage{color,soul}
\usepackage{amssymb}

\usepackage{bm}

\usepackage{subfigure}
\usepackage{algorithm}
\usepackage{algpseudocode}

\usepackage{hyperref}
\hypersetup{
    colorlinks=true,
    linkcolor=black,
    filecolor=black,      
    urlcolor=black,
    citecolor=black
}

\title{Correlative Channel-Aware Fusion for Multi-View Time Series Classification}
\author[1]{Yue Bai \thanks{bai.yue@northeastern.edu}}
\author[1]{Lichen Wang\thanks{wanglichenxj@gmail.com}}
\author[2]{Zhiqiang Tao\thanks{ztao@scu.edu}}
\author[3]{Sheng Li\thanks{sheng.li@uga.edu}}
\author[1]{Yun Fu\thanks{yunfu@ece.neu.edu}}
\affil[1]{Department of Electrical and Computer Engineering, Northeastern University, USA}
\affil[2]{Department of Computer Science and Engineering,
Santa Clara University, USA}
\affil[3]{Department of Computer Science, University of Georgia, USA}

\date{}

\begin{document}
\maketitle

\begin{abstract}
Multi-view time series classification (MVTSC) aims to improve the performance by fusing the distinctive temporal information from multiple views. Existing methods mainly focus on fusing multi-view information at an early stage, \emph{e.g.}, by learning a common feature subspace among multiple views. 
However, these early fusion methods may not fully exploit the unique temporal patterns of each view in complicated time series. 
Moreover, the label correlations of multiple views, which are critical to boosting, are usually under-explored for the MVTSC problem. 
To address the aforementioned issues, we propose a Correlative Channel-Aware Fusion (C$^2$AF) network.
First, C$^2$AF extracts comprehensive and robust temporal patterns by a two-stream structured encoder for each view, and captures the intra-view and inter-view label correlations with a graph-based correlation matrix. Second, a channel-aware learnable fusion mechanism is implemented through convolutional neural networks to further explore the global correlative patterns. These two steps are trained end-to-end in the proposed C$^2$AF network. Extensive experimental results on three real-world datasets demonstrate the superiority of our approach over the state-of-the-art methods. A detailed ablation study is also provided to show the effectiveness of each model component.

\end{abstract}

\section{Introduction}
Time series classification (TSC) has attracted much research attention recently, which provides more comprehensive information for the changing world.
Many algorithms are proposed for modeling time series data in different application domains, \emph{e.g.}, transportation~\citep{yao2018deep} and healthcare~\citep{harutyunyan2017multitask}. 
However, compared with static data such as images, the complicated dynamic patterns contained in time series make TSC a challenging problem.
Fortunately, owing to the advanced sensing techniques, objects or events can be observed through multiple modalities, which brings in multi-view time series data to improve the classification performance. For example, RGB, depth, and skeleton are three common modalities for human action recognition. They provide more comprehensive information to depict human actions than each single view. For another example, several types of human body signals are recorded as different modalities in health-care applications, such as magnetic resonance imaging (MRI) and electrocardiograph (ECG). These multi-view signals could monitor different physical states simultaneously. Generally, multi-view time series provide view-specific information from different angles and facilitate with each other for higher learning performance over an individual view.

\begin{figure}[t]
\centering
\begin{center}
\includegraphics[width=1.0\linewidth]{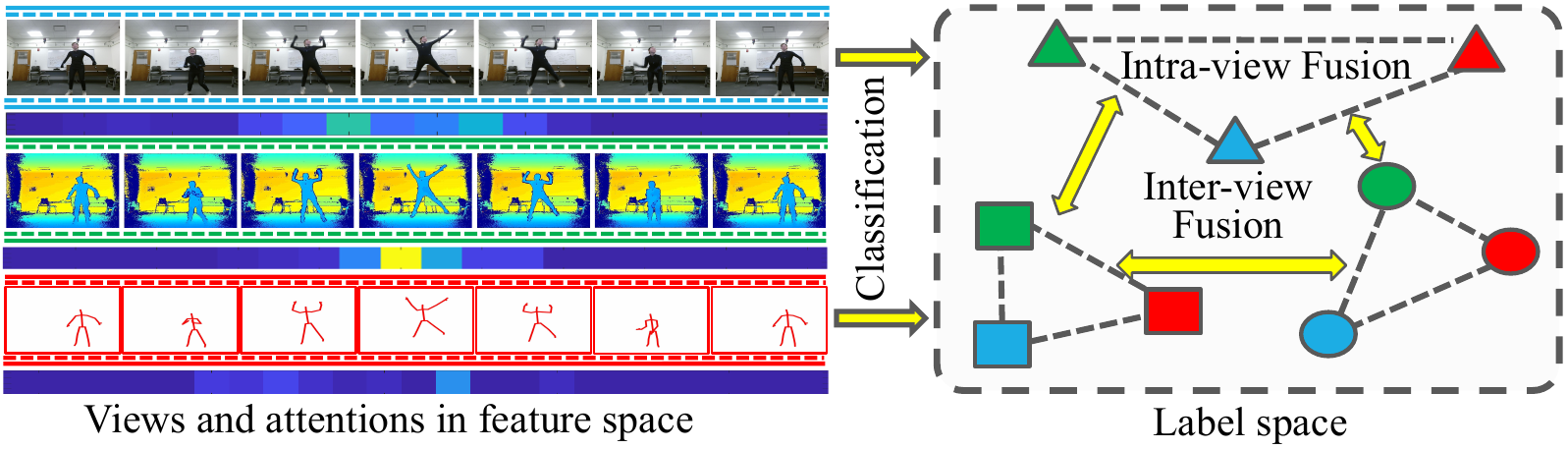}
\end{center}
\vspace{-5mm}
\caption{Multi-view temporal data has distinctive patterns in each view such as the attention scores. Intra-view/inter-view label correlations are crucial for boosting multi-view classification performance.
}
\label{fig:concept}
\vspace{-5mm}
\end{figure}


Multi-view learning (MVL) has drawn significant attention, since utilizing complementary information from different views has great potential to boost the final learning performance.
MVL is successfully applied in many applications~\citep{multiview_survey,AMGL_AAAI16,MLAN_AAAI17}. Previous algorithms could be roughly categorized into three groups~\citep{multiview_survey}: 1) co-training, 2) multiple kernel learning, and 3) subspace learning. Specifically, the co-training methods integrate multi-view data by maximizing the common mutual information of different views; the multiple kernel learning methods design specific learning kernels for each view and then combine them together; and the subspace learning methods seek for the common latent subspace shared by multiple views. Although these methods have achieved promising results, 
it is not straightforward to directly employ them for TSC due to the dynamic temporal patterns in time series.

Existing TSC methods focusing on single-view time series have been widely explored under
two cases: univariate~\citep{cuaresma2004forecasting} and multivariate~\citep{zheng2014time,husken2003recurrent}. On the one hand, the univariate TSC mainly studies the distance measurement between two time series such as~\citep{marteau2014recursive}. On the other hand, many research attempts are also made for handling the multivariate time series. To name a few, ~\citet{banko2012correlation} revised the dynamic temporal wrapping (DTW) method, and ~\citet{cui2016multi} utilized the convolutional neural networks (CNN) to model time series. Nevertheless, only a few methods are proposed for solving multi-view and multivariate TSC. For instance, ~\citet{li2016multi} proposed a discriminative bilinear projection framework to build a shared latent subspace for multi-view temporal data. \citet{zadeh2018memory} designed a fusion strategy based on long short-term memory (LSTM) networks. \citet{yuan2018muvan} proposed an attention mechanism to model multi-view time series. It is worth noting that, all these methods adopt an early fusion strategy, \emph{e.g.}, integrating multi-view information by learning a common feature subspace, which may not fully exploit the view-specific distinctive patterns and ignore the multi-view label correlations. 

To address the above issues, we propose a Correlative Channel-Aware Fusion (C$^2$AF) network for the multi-view time series classification (MVTSC) task. Our C$^2$AF jointly leverages the view-specific distinctive temporal patterns existing in feature spaces and the multi-view correlations in label spaces (see Figure~\ref{fig:concept}), to boost the classification performance. Specifically, our model first applies a two-stream temporal encoder to extract robust temporal features, followed by a classifier for each view. By this means, the raw label information is first obtained. After that, the multi-view label correlations are captured by a graph-based correlation matrix. Finally, a channel-aware learnable fusion mechanism is designed to globally integrate the label correlations and tune the entire network. The main contributions of our work are summarized as follows:
\begin{itemize}
\item We propose an end-to-end MVTSC network, namely C$^2$AF, to jointly capture view-specific temporal patterns by two-stream encoders and automatically fuse the multi-view label correlations. 

\item We design a channel-aware learnable fusion mechanism, which provides an effective late fusion strategy for the MVTSC problem and adopts a concise implementation via convolutional neural networks.


\item We conduct extensive experiments on three real-world datasets to show the superiority of C$^2$AF over state-of-the-art methods, and provide detailed ablation studies to demonstrate the effectiveness of each model component.

\end{itemize}

\begin{figure*}[t]
\centering
\begin{center}
\includegraphics[width=1.0\linewidth,height=45mm]{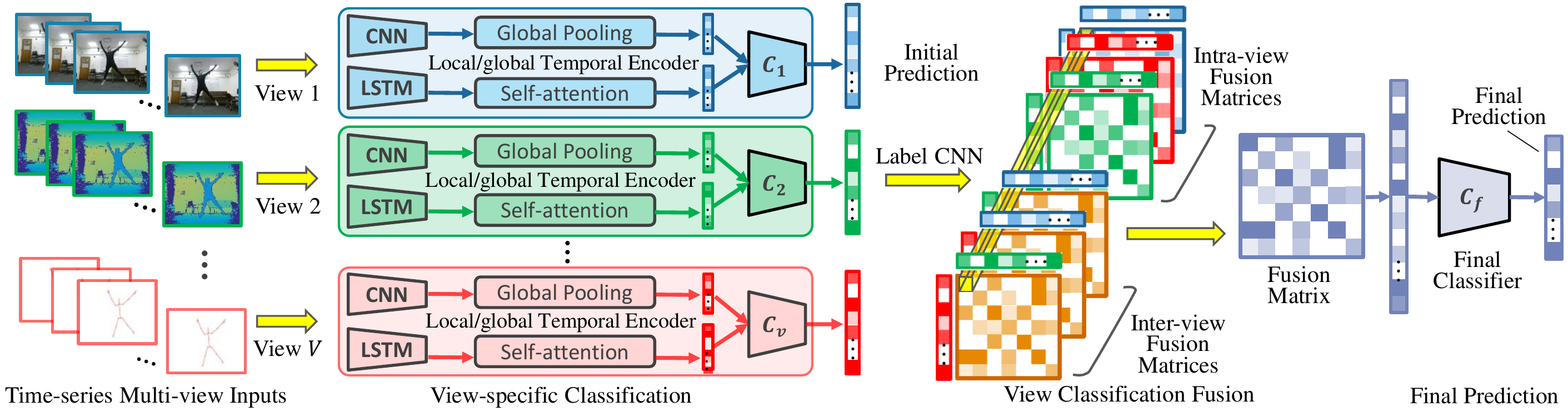}
\end{center}
\vspace{-2mm}
\caption{Multi-view temporal data are set as input simultaneously to train the end-to-end C$^2$AF network. A two-stream encoder extracts view-specific temporal patterns. Intra-view/inter-view label correlations are captured by correlation matrices. The channel-aware learnable fusion integrates and fully utilizes multi-view label correlations for performance improvement.}
\label{fig:framework}
\vspace{-2mm}
\end{figure*}
\section{Related Work}\label{sec:related_work}

\subsection{Time Series Classification}
Time series data are collected and analyzed in a wide range of domains~\citep{xing2010brief}. Generally, the methods focusing on time series classification (TSC) task can be categorized into three groups: 1) feature based classification, 2) sequence distance based classification, and 3) model based classification. Feature based algorithms such as~\citep{kadous2005classification, ye2009time} extract a feature vector from time series and then apply traditional methods \emph{e.g.}, support vector machine (SVM)~\citep{cortes1995support} and K-Nearest neighbor (KNN)~\citep{fukunaga1975branch} to make classification. Further, deep neural network has great capacity to fit non-linear mapping and extract complicated temporal features for classification~\citep{karim2019multivariate}. Reservoir computing~\citep{bianchi2018reservoir} is proposed based on recurrent neural networks to learn the representations for multivariate TSC. Distance based methods aim to design distance functions to measure the similarity of a pair of time series. As long as obtaining a reasonable distance metric, we apply conventional algorithms to further make classification. For example, DTW~\citep{xi2006fast} is a typical distance based algorithm which is eligible for time series with different lengths. Other distance based models are also proposed for TSC such as~\citep{wei2006semi,ratanamahatana2004making, keogh2003need}. Model based methods assume that all time series belonging to each class are generated by a potential generative model. During the training stage, the corresponding parameters of the potential model are learned and the test samples are classified based on the likelihood. To name a few, hidden markov model (HMM)~\citep{rabiner1989tutorial} is widely used in TSC for speech recognition. Naive bayes sequence classifier~\citep{rish2001empirical} is another typical model based method which observes the feature independent assumption. In our work, we mainly focus on multi-view time series classification (MVTSC) problem which is not fully explored by above methods.

\subsection{Multi-View Learning}
Multi-view learning (MVL) attracts increasing attention in recent decades. The distinctive patterns extracted from different views mutually support with each other to benefit final performance. MVL is widely used in many tasks such as object classification~\citep{qi2016volumetric}, clustering~\citep{bickel2004multi}, semi-supervised learning~\citep{hou2010multiple}, action recognition~\citep{cai2014multi}, and face recognition~\citep{li2002statistical}. Fusing information from multiple views is an effective way to leverage mutual-support patterns for performance improvement in MVL~\citep{swoger2007multi, bruno2009multiview}. Fusion strategies can be categorized into three groups~\citep{atrey2010multimodal}: 1) feature fusion, 2) decision fusion, and 3) hybrid fusion. Feature fusion (early fusion)~\citep{wang2017select, louis2011towards, poria2016data} focuses on merge distinctive information from different views in feature space. Decision fusion (late fusion)~\citep{wortwein2017really, nakano2016icmi} aims to fuse the multiple decisions in label space. Hybrid fusion is a combination of feature fusion and decision fusion. However, most existing MVL methods are not designed for temporal data. Deploying them on time series directly will ignore temporal dynamic patterns. 
In our work, we propose a novel Correlative Channel-Aware Fusion (C$^2$AF) network for MVTSC. 
Our proposed C$^2$AF extracts robust temporal representations and fully explores the multi-view latent correlations.

\section{Methodology}\label{sec:method}

\subsection{Preliminary}
Let $\mathcal{X} = \{X^{v}\}_{v=1}^{V}$ be the multi-view time series data, where $X^{v} \in \mathbb{R}^{T \times D^{v}}$ refers to the $v$-th view feature matrix. For $\forall v$, $T$ and $D^{v}$ represent the time series length and feature dimensions, respectively. Let $Y \in \mathbb{R}^{K}$ be the corresponding label, where $K$ denotes the number of classes. All the views in $\mathcal{X}$ share the same label $Y$. In this study, we focus on multi-view time series classification (MVTSC) by leveraging multi-view complementary information through a Correlative Channel-Aware Fusion (C$^2$AF) network. Our C$^2$AF can be divided into two parts, global-local temporal encoder and channel-aware learnable fusion.

\subsection{Global-Local Temporal Encoder}
Dynamic and complicated temporal pattern is the key factor to tackle time series data. It usually provides discriminative characteristics to guarantee high quality classification. In our C$^2$AF approach, obtaining comprehensive and robust temporal representations for each view is indispensable, which provides reliable label information and benefits fusion process. We propose a global-local temporal encoder to fully explore the temporal context. It consists of a global-temporal encoder $E_g$ and a local-temporal encoder $E_l$. We obtain view specific representations by 
\begin{equation}\label{eq:H-enc}
\begin{aligned}
    	H^v & = q(H_g^v, H_l^v)\\
    H_{g}^v & = E_g(X^v; \phi^{v}_{g})\\
    H_{l}^v & = E_l(X^v; \phi^{v}_{l}),
\end{aligned}
\end{equation}
where $H^v \in \mathbb{R}^{d^{v}}$ is the encoded representations for $X^v$, $H_{g}^v/H_{l}^v$ represents the $E_g/E_l$ output, $q$ denotes a common fusion operation (we use concatenation operation in our work), and $E_g$, $E_l$ are two networks with learnable parameters $\phi^{v}_{g}$ and $\phi^{v}_{l}$, respectively. We update $\phi^{v}_{g}$ and $\phi^{v}_{l}$ by minimizing the following loss:
\begin{equation}\label{eq:Lv}
L^v = \sum_{i=1}^{N} \ell(Y_i, \hat{Y}_i^{v}),
\end{equation}
where $\ell$ represents the cross-entropy loss and $N$ is the number of samples. $\hat{Y}_i^{v} = C_v(H^{v}_i)$ is the prediction for the $i$-th sample. $C_v: \mathbb{R}^{d^{v}} \rightarrow \mathbb{R}^{K}$ is the $v$-th view specific  classifier achieved by a linear mapping.

\subsubsection{Global-Temporal Encoder}

Next, we will introduce $E_g$ and $E_l$ with more details. The $E_g$ and $E_l$ are deployed for each view. For convenience, we omit the subscript $v$ in the rest of this section.
We adopt recurrent neural networks (RNN) to parameterize our global-temporal encoder $E_g$, as RNN has been well validated as an effective way to explore the temporal context for time-series. Particularly, we employ the long short-term memory (LSTM)~\citep{hochreiter1997long} as the RNN cell, which is given by 
\begin{equation}\label{eq:LSTM}
\begin{aligned}
    f_t & = \sigma_{g}(W_f x_t + U_f h_{t-1} + b_f), \\
    i_t & = \sigma_{g}(W_i x_t + U_i h_{t-1} + b_i), \\
    o_t & = \sigma_{g}(W_o x_t + U_o h_{t-1} + b_o), \\ 
    c_t & = f_t \circ c_{t-1} + i_t \circ \sigma_c (W_c h_t + U_c h_{t-1} + b_c), \\
    h_t & = o_t \circ \sigma_h(c_t),
 \end{aligned}
\end{equation}
where $x_t$ is $t$-th representation in sequence input $X$ ($1 \leq t \leq T$). $f_t$, $i_t$, $o_t$, $c_t$, and $h_t$ denote forget gate, input gate, output gate, cell state, and hidden state at time $t$, respectively. $c_{t-1}$ and $h_{t-1}$ are cell and hidden states at time $t-1$. $\sigma_g$, $\sigma_c$, $\sigma_h$ are activation functions, and $\circ$ represents the element-wise product. In Eq.\eqref{eq:LSTM}, $W_{*}$, $U_{*}$ and $b_{*}$ are all learnable weights, $\forall * \in \{f, i, o, c\}$.

To further enhance the global temporal representation, we leverage attention mechanism to integrate the hidden states sequence. By using attention, we explicitly learn the dynamic correlations cross different time points, and obtain the global temporal representation $H_{g}$ by 
\begin{equation}\label{eq:Att}
    H_{g} = \sum_{t=1}^{T} \omega_t h_t,
\end{equation}
where $\omega = \{\omega_t\}$ is the learnable attention weights. 

By using Eqs.~(\ref{eq:LSTM}-\ref{eq:Att}), we formulate our $E_g$ as LSTM with attention mechanism, and have $\phi_{g} = \{\{W_{*}, U_{*}, b_{*}\}, \omega\}$, $* \in \{f, i, o, c\}$.
\begin{algorithm}[tb]
  \caption{The procedure of training C$^2$AF algorithm.}
  \label{alg:CCF}
  \begin{algorithmic}[1]
    \Require
        batches of $\{\mathcal{X}, Y\}$, number of view $V$, 
        number of training steps $S$
    \Ensure
        prediction of each view $\hat{Y}^{v}$ and final result $\hat{Y}^{f}$
    \For{each $i \in [1,S]$}
    \For{each $v \in [1,V]$}
    \State sample a batch data $X^v$ from view $v$
    \State forward $X^v$ into $E_g$ and $E_l$
    \State compute $H^v$ and $\hat{Y}^v$ through Eq.~(\ref{eq:H-enc}) and $C_v$
    \State update $\phi_g^v$, $\phi_l^v$ and $C_v$ using Eq.~(\ref{eq:Lv}) 
    \EndFor
    \State forward $\hat{Y}^v, v\in {1,2,...,V}$ into $E_f$
    \State compute $\hat{Y}^f$ through Eq.~(\ref{eq:fusion}) and $C_f$
    \State update $\phi_f$ and $C_f$ using Eq.~(\ref{eq:Lf}) 
    \EndFor
\State\Return $\hat{Y}^{v}$ and $\hat{Y}^{f}$
  \end{algorithmic}
\end{algorithm}
\subsubsection{Local-Temporal Encoder}
Different from the global-temporal encoder, we utilize convolutional neural networks (CNN) to formulate our local-temporal encoder $E_l$, as CNN works well on probing patterns from local-characterized data. Specifically, we apply a set of 1D convolutional filters to extract local patterns in $X$ following the similar strategy in temporal convolutional networks (TCN)~\citep{lea2016temporal}. Let $M$ be the number of CNN layers and $F_m \in \mathbb{R}^{T_m \times D_m}$ be the output of the $m$-th layer ($1 \leq m \leq M$). $T_m$ and $D_m$ denote the corresponding temporal and feature dimensions, respectively. Given $F_0 = X$, we compute $F_m$ by
\begin{equation}\label{eq:CNN}
\begin{aligned}
    F_m = \textup{BN}_{\{\gamma_m, \beta_m\}}(\textup{ReLU}(
    W_m * F_{m-1} + b_m)),
 \end{aligned}
\end{equation}
where $W_m \in \mathbb{R}^{D_m \times D_{m-1} \times \Delta T}$ is the weight of convolutional filter, $b_m \in \mathbb{R}^{D_m}$ is the bias. $\Delta T$ represents the size of temporal sliding window and $*$ denotes the convolution operation. In Eq.~\eqref{eq:CNN}, $\textup{BN}_{\{\gamma_m, \beta_m\}}$ refers to the batch normalization block~\citep{ioffe2015batch} with learnable parameters $\gamma_m$ and $\beta_m$. It is used to further improve the effectiveness and stability of $E_l$.

In order to reduce the number of parameters and avoid over-fitting issue, a global average pooling layer~\citep{lin2013network} is deployed after each convolutional block. By using these methods, we efficiently extract local temporal information and obtain high-level representation $H_l$ by
\begin{equation}\label{eq:pool}
\begin{aligned}
    H_l = g(F_M),
 \end{aligned}
\end{equation}
where $g$ is the global average pooling layer.

Through applying Eqs.~(\ref{eq:CNN}-\ref{eq:pool}), we concretize our $E_l$ as CNN with batch normalization and global average pooling, and have $\phi_{l} = \{W_m, b_m, \gamma_m, \beta_m\}_{m=1}^{M}$.

\subsection{Channel-Aware Learnable Fusion}
Efficiently fusing mutual-support information from multi-view predicted labels $\hat{Y}^{v}$ ($1 \leq v \leq V$) is the central fact of performance improvement. In our model, we propose a channel-aware learnable fusion mechanism to sufficiently capture and utilize the label correlations. It takes advantage of intra-view and inter-view label correlations to achieve better multi-view learning results. Specifically, we construct a graph based correlation matrix to probe intra-view and inter-view label correlations and a CNN based fusion module to integrate global patterns. Next, we introduce the channel-aware learnable fusion with more details.

\subsubsection{Label Correlation Matrix}\label{sec:LCM}
We adopt a graph based strategy to capture the intra-view and inter-view label correlations, respectively. The intra-view label correlation matrix for each view $v$ is given by
\begin{equation}\label{eq:intra}
G^{v,v} = \hat{Y}^v \cdot \hat{Y}^{v\top},
\end{equation}
where $G^{v,v} \in \mathbb{R}^{K \times K}$ is the correlation matrix derived by multiplying the predicted label $\hat{Y}^v \in \mathbb{R}^{K \times 1}$ and its transpose $\hat{Y}^{v\top} \in \mathbb{R}^{1 \times K}$ for $1 \leq v \leq V$. Each element in $G^{v,v}$ represents the intra-view pair-wise label correlations for view $v$. We integrate $V$ intra-view label correlations by concatenating them together as follow:

\begin{equation}\label{eq:intra_concat}
r_{intra} = [G^{1,1}, G^{2,2},..., G^{V,V}],
\end{equation}
where $r_{intra} \in \mathbb{R}^{K \times K \times V}$ is the intra-view correlation tensor and $[\cdot]$ is the concatenation operation. 

Similarly, the inter-view label correlation matrix for each pair of views is given by
\begin{equation}\label{eq:inter}
G^{u,w} = \hat{Y}^u \cdot \hat{Y}^{w\top},
\end{equation}
where $G^{u,w} \in \mathbb{R}^{K \times K}$ is the correlation matrix derived by multiplying the predicted label $\hat{Y}^u \in \mathbb{R}^{K \times 1}$ from view $u$ and the transpose of predicted label $\hat{Y}^w \in \mathbb{R}^{1 \times K}$ from view $w$ for $\forall u, w \in V, u \neq w$. Each element in $G^{u,w}$ represents the inter-view pair-wise label correlations for view $u$ and $w$. Considering all the possible combinations of view-pair, we integrate $\binom{V}{2}$ inter-view label correlations by concatenating them together as follow:

\begin{equation}\label{eq:inter_concat}
r_{inter} = [G^{1,2}, G^{1,3},..., G^{V-1,V}],
\end{equation}
where $r_{inter} \in \mathbb{R}^{K \times K \times \binom{V}{2}}$ represents the inter-view correlation tensor. 

By using Eqs.~(\ref{eq:intra}-\ref{eq:intra_concat}) and Eqs.~(\ref{eq:inter}-\ref{eq:inter_concat}), we extract the intra-view and inter-view label correlations as two multi-channel tensors $r_{intra}$ and $r_{inter}$.


\subsubsection{Channel-Aware Fusion}
Multi-view label correlations are extracted and represented by label correlation matrices. The informative patterns of label correlations are reserved in each element instead of a local area of these matrices, but still contained in the same place across different channels of $r_{intra}$ and $r_{inter}$. Hence, we employ a CNN structure with $1 \times 1$ kernels as a
channel-aware extractor to globally integrate cross-view correlative information. It is given by

\begin{equation}\label{eq:fusion}
\begin{aligned}
r = E_f([r_{intra}, r_{inter}], \phi_f),
\end{aligned}
\end{equation}
where $r \in \mathbb{R}^{K \times K \times N_{k}}$ is the fusion matrix. 
$E_f$ is the CNN based fusion encoder parameterized by $\phi_f$, with $N_{k}$ kernels. We formulize the fusion encoder $E_f$ by

\begin{equation}\label{eq:late}
r_{p,q}^{(o)} = f(b^{(o)} + \langle W^{(o)}, [r_{intra}, r_{inter}]_{p,q}\rangle),
\end{equation}
where $r_{p,q}^{(o)}$ is the $(p,q)$ element of $r^{(o)} \in \mathbb{R}^{K \times K \times 1}$ which is the $o$-th component of $r$ for $1 \leq o \leq N_k$. $W^{(o)} \in \mathbb{R}^{1 \times 1 \times (V + \binom{V}{2})}$ and $b^{(o)} \in \mathbb{R}^{1 \times 1}$ are the learnable weights and bias of $ 1 \times 1 $ filter. $[r_{intra}, r_{inter}]_{p,q}$ represents the $(p,q)$ element of cross-view correlation tensor concatenated by $r_{intra}$ and $r_{inter}$. $f$ is the activation function.

Through Eqs.~(\ref{eq:fusion}-\ref{eq:late}), we formulate our fusion encoder $E_f$, and have $\phi_{f} = \{W, b\}$. We update $\phi_f$ by minimizing the following loss:
\begin{equation}\label{eq:Lf}
L^f = \sum_{i=1}^{N} \ell(Y_i, \hat{Y}_i^{f}),
\end{equation}
where $\hat{Y}_i^{f} = C_f(T_{flatten}(r_{i}))$ is the prediction for the $i$-th sample.
$T_{flatten}$ is a flatten operation to transfer feature matrix $r_{i}$ into a vector, and $C_f: \mathbb{R}^{D_f} \rightarrow \mathbb{R}^{K}$ is the final classifier achieved by a linear mapping with $D_f = K \times K \times N_k$. During the training, we alternatively optimize the set of loss $L^v$ for each view and $L^f$ for the final classifier. The entire procedure of training our proposed C$^2$AF algorithm is summarized in Algorithm~\ref{alg:CCF}.

\begin{table*}[t]
\setlength{\tabcolsep}{4pt}
\centering
\caption{Classification performance on three datasets}\label{tab:main}
\scalebox{0.85}{
\begin{tabular}{ c  c  c  c  c  c  c  c  c  c  c  c}
\toprule
\multirow{2}{*}{Dataset} & \multicolumn{4}{c}{EV-Action} & \multicolumn{4}{c}{NTU RGB+D} & \multicolumn{3}{c}{UCI}\\
\cmidrule(lr){2-5}
\cmidrule(lr){6-9}
\cmidrule(lr){10-12}
& RGB & Depth & Skeleton & Three-view & RGB & Depth & Skeleton & Three-view & View1 & View2 & Two-view\\
\midrule
MFN & 0.5752 & 0.3978 & 0.6603 & 0.4769 &0.6830&0.7630&0.6854&0.8159 & 0.5563 & 0.7141 & 0.7260\\
RC classifier & 0.5990 & 0.5790 & 0.7850 & 0.6130 &0.7829&0.8013&0.6765&0.8270& 0.7660 & 0.7700 & 0.8190\\
MLSTM-FCN & 0.6814 & 0.6914 & 0.7613 & 0.7555 &0.7760&0.7929&0.6778&0.8284& 0.8754 & 0.9246 & 0.9208\\
\midrule
Concat-LSTM & - & - & - & 0.7325 &-&-&-&0.8330& - & - & 0.8290\\
Concat-CNN & - & - & - & 0.6132 &-&-&-&0.8295& - & - & 0.8919\\
Label-Concat & 0.7124 & 0.7134 & 0.7585 & 0.8206 &0.7304&0.8113&0.7235&0.8402& 0.8643 & 0.8535 & 0.9090\\
Label-Average & 0.7285 & 0.7114 & 0.7505 & 0.8156 &0.7214&0.8090&0.7402&0.8319& 0.8699 & 0.8559 & 0.8728\\
Label-Max & 0.7575 & 0.7044 & 0.7615 & 0.8026 &0.7239&0.8006&0.7287&0.8221& 0.8704 & 0.9052 & 0.9113\\
C$^2$AF (Ours) & 0.7615 & 0.7284 & 0.7645 & \textbf{0.8406} &0.7248&0.8034&0.7347&\textbf{0.8688}& 0.8656 & 0.9027 & \textbf{0.9314}\\
\bottomrule
\end{tabular}}
\vspace{-4mm}
\end{table*}

\section{Experiments}
\subsection{Experimental Setting}
\subsubsection{Datasets}\label{sec:dataset}
We utilize three real-world multi-view time series datasets to prove the model effectiveness.
\begin{itemize}

    \item \textbf{EV-Action}~\citep{wang2019ev} is a multi-view human action dataset. We choose RGB, depth, and skeleton views for our multi-view time series experiments. EV-Action contains 20 human common actions and 53 subjects performing each action 5 times, so that we have 5300 samples in total. We choose the first 40 subjects as training set and the rest 13 subjects as test set.


    \item \textbf{NTU RGB+D}~\citep{shahroudy2016ntu} is a large-scale multi-view dataset for action recognition. It contains 56000 action clips in 60 action classes performed by 40 subjects. We choose the RGB, depth, and skeleton views for our experiments. We use the cross-subject benchmark provided by the original dataset paper, which contains 40320 samples for training and 16560 samples for test.
    \item \textbf{UCI Daily and Sports Activities}~\citep{asuncion2007uci} is a multivariate time series dataset, which contains motion sensor data of 19 human daily actions. There are 45 sensors placed on subject's body. Each activity is performed by 8 subjects and has 480 samples. We follow the same mutli-view experimental setting from~\citep{li2016multi} in our model evaluation.
\end{itemize}


\subsubsection{Baseline Methods}

Several baseline methods including the state-of-the-art approaches are deployed to demonstrate our model effectiveness. 
\begin{itemize}

    \item \textbf{MLSTM-FCN}~\citep{karim2019multivariate} is a novel deep framework proposed to handle multivariate time series data, which achieves promising performances on extensive real-world time series datasets.

    \item \textbf{RC Classifier}~\citep{bianchi2018reservoir} proposes a reservoir computing (RC) framework to encode multivariate time series data as a vectorial representation in an unsupervised fashion, which has a relatively low computational cost during handling temporal data.
    \item \textbf{MFN}~\citep{zadeh2018memory} designs a memory fusion mechanism as an early fusion approach to tackle with multi-view time series.
    \item \textbf{Concat-LSTM/Concat-CNN} fuses multi-view time series using concatenation operation as input for LSTM and CNN. We use them as two early fusion baselines.
    \item \textbf{Label-Concat/Label-Average/Label-Max} fuses the predicted labels from multiple views using concatenation, average pooling, and max pooling, respectively. We use them as three late fusion baselines. 
\end{itemize}
To adopt MLSTM-FCN and RC classifier for MVTSC, we concatenate multi-view time series as a multivariate time series along with the feature dimension as model input.
MFN is designed for multi-view learning, we use it directly for model evaluation. 
We report the single-view and multi-view performances simultaneously for comparison except Concat-LSTM and Concat-CNN as they cannot provide single-view output.


\begin{figure}[t]
\centering
\subfigure[]
{
   \label{fig:subfig:a} 
   \includegraphics[height=2.0in]{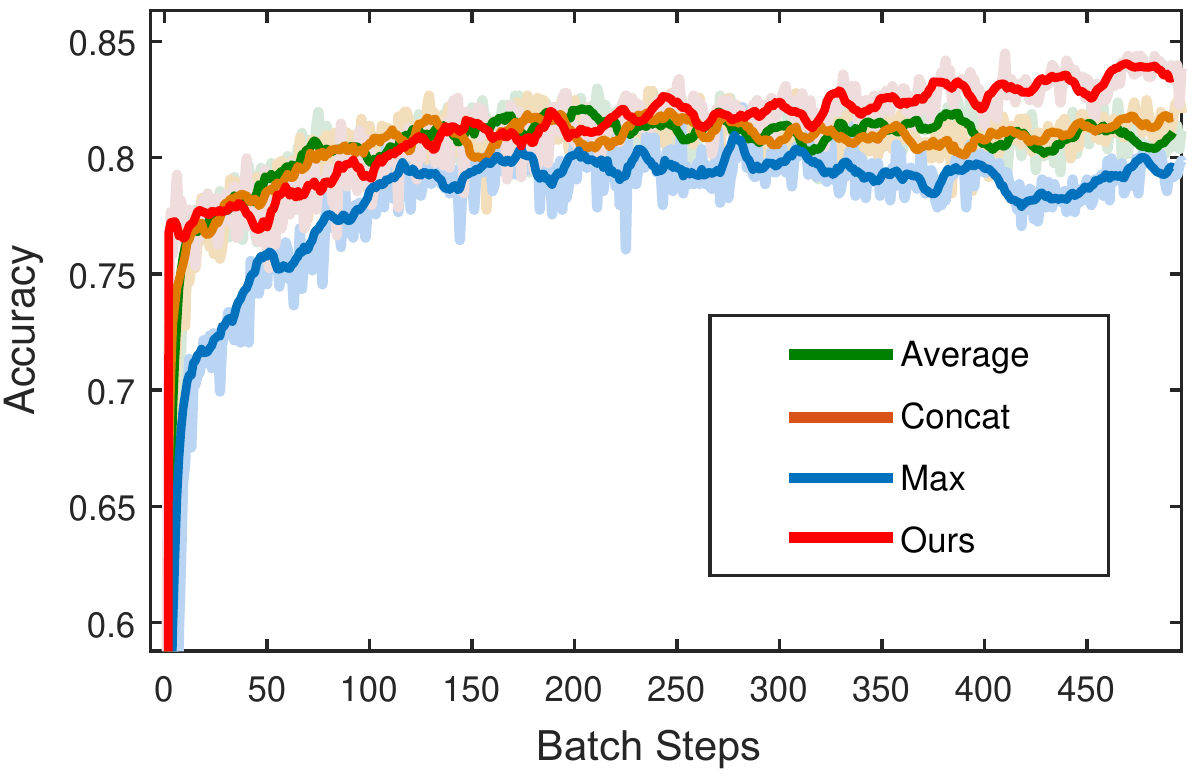}\label{fig:baseline}
}
\subfigure[]
{
\label{fig:subfig:b} 
\includegraphics[height=2.0in]{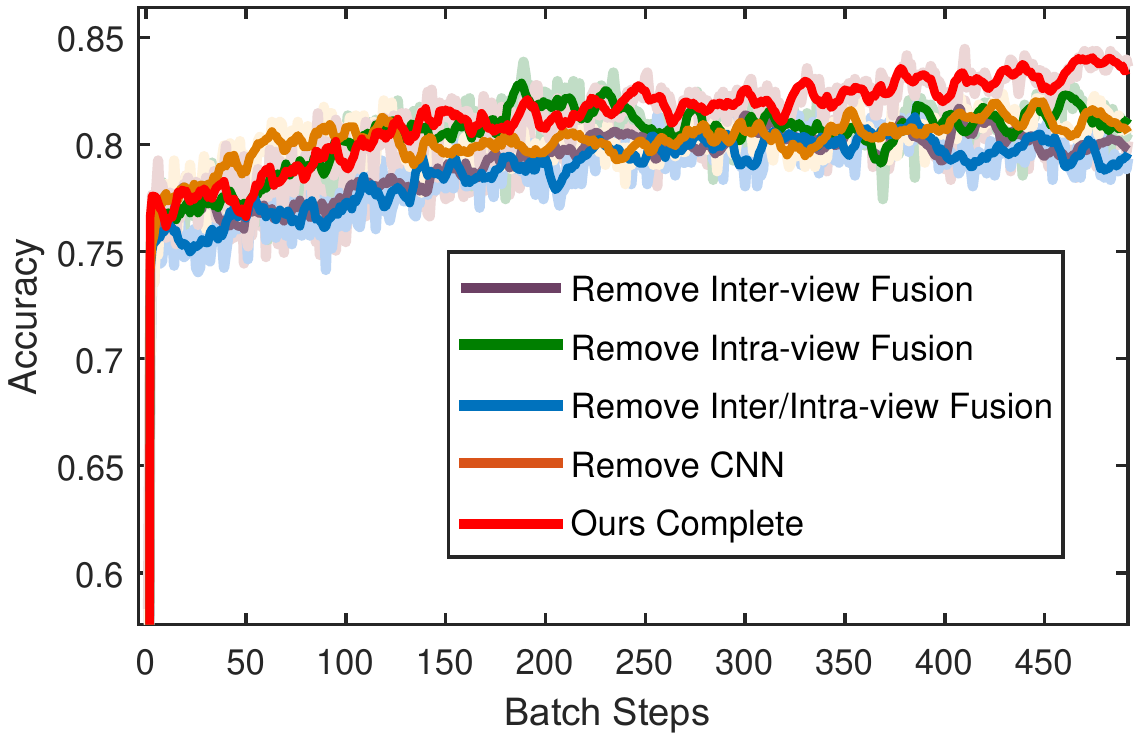}\label{fig:ablation}
}
\caption{(a) Comparisons between our model and late fusion baselines which prove that our channel-aware fusion is an effective and efficient fusion strategy. Shadow lines denote the exact performances per batch step, while the solid lines indicate the smoothed performances. (b) Ablation study on channel-aware learnable fusion. Shadow lines denote the exact performances per batch step, the solid lines indicate the smoothed performances.  }
\label{fig:subfig} 
\end{figure}

\subsubsection{Data Preprocessing}
We utilize the same strategy to preprocess multi-view data for EV-Action and NTU RGB+D as they both have RGB, depth, and skeleton views. Specifically, we align all the samples into the same 60 length with cutting and repeating strategies for longer and shorter samples. Next, we adopt TSN~\citep{wang2016temporal} to extract frame-level features for RGB view with pre-trained BNIncepction backbone. The depth view is transferred into RGB format firstly using HHA algorithm~\citep{gupta2014learning} and extract fetures using exactly the same strategy as RGB view. For skeleton view, we concatenate 3D coordinates of 25 joints at each time point as frame-level features. Specifically, in order to easily handle the large-scale skeleton data in NTU RGB+D dataset, we use VA-LSTM~\citep{zhang2017view} as the backbone to preprocess the 3D coordinates data.
As a summary, for EV-Action and NTU RGB+D datasets, RGB, depth, and skeleton data are extracted as frame-level features with 60 temporal length and 1024, 1024, and 75 feature dimensions, respectively.



We follow the same data preprocessing procedure in~\citep{li2016multi} for UCI Daily Sports dataset. As a summary, the sensor data are set as View1 and View 2 with 125 temporal length, 27 and 18 feature dimensions, respectively.


\subsection{Implementation}
As shown in Figure~\ref{fig:framework}, the frame-level features of each view are set as input of global-local temporal encoder simultaneously to obtain the view-specific representations. The outputs of global-temporal encoder and local-temporal encoder are concatenated as the input of view-specific classifier $C_v$. Each $C_v$ is trained by optimizing its corresponding loss $L_v$. The predicted label from different views $\hat{Y}^v$ construct two sets of correlation matrices for capturing intra-view and inter-view label correlations. The cross-view correlative tensor is derived by stacking all the correlation matrices and fed into channel-aware learnable fusion module. Fused feature vector is set as input to train classifier $C_f$ for final prediction through optimizing $L^f$. We set the batch size as 128. The Adam optimizer~\citep{kingma2014adam} is applied for optimization and the learning rates are set as 0.0001 for all the view-specific and final classifiers synchronously. During the training process, the classifiers of all views $C_v$ are trained firstly to obtain the initial classification results which makes a concrete foundation for the learnable fusion module. Next, the final classifier $C_{f}$ is trained based on the initial predicted labels. After that, $C_v$ and $C_{f}$ are trained alternatively during the whole training process and we report the single-view and final performances simultaneously. Our model is implemented using Tensorflow with GPU acceleration.

\subsection{Performance Analysis}
Classification performances for three datasets are shown in Table~\ref{tab:main}. For EV-Action dataset, RGB, Depth and Skeleton represent the single-view classification accuracy, while the Three-view indicates the multi-view classification results. The skeleton view is the most informative view which always achieves the best performance compared with RGB and depth views. The baseline methods obtain comparable even better performances on single-view, however, our proposed model achieves the best performance for multi-view results and outperforms each single-view result. MFN cannot make early fusion efficiently to improve multi-view performance on EV-Action dataset which indicates the early fusion of MFN is not capable of handling high dimensional temporal data. Besides, the large difference between the feature dimensions of different views also hinders the memory fusion process resulting in low performances. However, our C$^2$AF will not suffer from this issue since we focus on extracting label correlations for multi-view fusion. RC classifier and MLSTM-FCN achieve competitive results on skeleton view, however they cannot effectively fuse multi-view information for better performance. The comparisons with three simple late fusion methods prove our learnable fusion is a more effective fusion strategy. We visualize the comparisons between late fusion baselines and our C$^2$AF in Figure~\ref{fig:baseline}, which shows the performance variations along with batch steps.

For NTU RGB+D dataset, all the baselines perform much better results compared with EV-Action, since the NTU RGB+D is a very large-scale dataset which provides sufficient training samples. The depth is the most informative view. All the comparison methods take advantage of the sufficient data to achieve better results on multi-view scenario. However, our C$^2$AF fully explores the multi-view latent correlations and still achieves the best MVTSC performance.

For UCI dataset, View1 and View2 represent the two single-view and Two-view indicates the multi-view performances. View2 always obtains better results for single-view compared with View1. The baseline methods achieve competitive results for single-view but cannot outperform our fusion strategy. MFN improves the multi-view performance compared with single-view, however, it is still lower than our model. 
MLSTM-FCN obtains high performance for both single-view and multi-view, however, it cannot utilize multi-view data sufficiently for further improvement. Our proposed model achieves the best multi-view performance.

\begin{table}[t]
\setlength{\tabcolsep}{4pt}
\centering
\caption{Global/Local Encoder}\label{tab:twostream}
\scalebox{1.0}{
\begin{tabular}{ c  c  c  c  c  c}
\toprule
Dataset & \multicolumn{3}{c}{EV-Action} & \multicolumn{2}{c}{UCI} \\

\ View & RGB & Depth & Skeleton & View1 & View2 \\
\midrule
Local Encoder & 0.6263 & 0.6192 & 0.7735 & 0.8730 & 0.9001 \\
Global Encoder & 0.7104 & 0.7084 & 0.7665 & 0.7194 & 0.8292 \\
\bottomrule
\end{tabular}
}
\vspace{-4mm}
\end{table}

\subsection{Model Analysis}
We design detailed ablation study to prove the necessity of each model component. First, we use global temporal encoder and local temporal encoder individually to make view-specific classification on two datasets as shown in Table~\ref{tab:twostream}. For EV-Action dataset, global encoder works better than local encoder. However, for UCI dataset, the local encoder outperforms global encoder for both two views. Hence, our two-stream temporal encoder is indispensable to handle diverse time series data. It takes advantage of global and local encoders to obtain robust temporal representations.

We divide our learnable fusion module into several parts to make ablations.
The whole fusion module can be separated as two parts, label correlative matrix and channel-aware fusion. Further, the label correlative matrix can be divided into intra-view and inter-view parts. The experimental results are shown in Table~\ref{tab:ablation}. \textbf{Intra-view Only}/\textbf{Inter-view Only} represents we only use intra-view/inter-view matrices. 
\textbf{Channel-aware Fusion Only} means we remove all the correlative matrices and concatenate predicted label vectors together as input to channel-aware fusion which proves the necessity of our whole correlative matrices. \textbf{Ours without Channel-aware fusion} indicates that we directly flatten all the correlation matrices into one feature vector as input to final classifier. 
The results illustrate each model component cannot obtain the best performance individually, while the complete C$^2$AF achieves the best accuracy. We visualize the performance curves of ablation in Figure~\ref{fig:ablation}, which shows the performance variations along with batch steps.

\begin{figure*}[t]
\centering
\begin{center}
\includegraphics[width=1.0\linewidth]{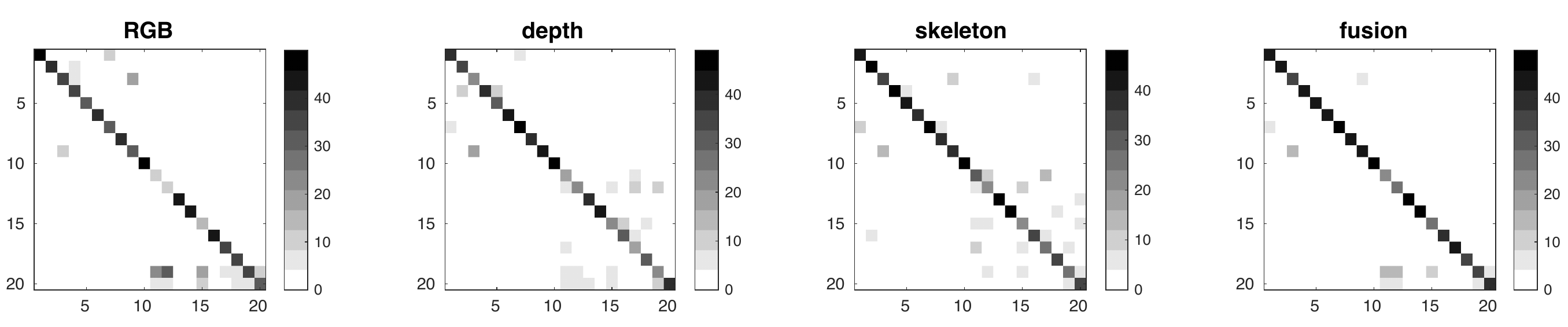}
\end{center}
\vspace{-1mm}
\caption{Confusion matrices for each single view and multi-view fusion on EV-Action dataset. The first 10 classes are performed by subjects themselves (\emph{e.g.}, standing, walking, and jumping). The last 10 classes are performed with interactive objects (\emph{e.g.}, moving table, reading book, and throwing ball). The colorbar denotes the relationship between the correct predictions and color. The deeper color represents the better classification performance.
}
\label{fig:confusion}
\vspace{-2mm}
\end{figure*}

To better understand how the learnable fusion process benefits the MVTSC, we show the classification confusion matrices for each single view and multi-view fusion on EV-Action in Figure~\ref{fig:confusion}. In EV-Action, classes can be divided into two groups~\citep{wang2019ev}: the first 10 actions are performed by subjects themselves, and the last 10 actions are performed interactively with other objects. RGB and depth views can accurately distinguish if the action is interactive, but easily make mistakes within each group. Skeleton view 
is not sensitive to the interactive objects so that it still makes mistakes cross these two groups. But its results are generally better than RGB and depth. We observe that our method takes full advantages of different views: it fuses the patterns from RGB and depth views to distinguish accurately if the action is interactive; it also benefits from the skeleton view to reduce the mistakes occurred within each group, so that our C$^2$AF achieves the most reasonable results.

\begin{table}[t]
\setlength{\tabcolsep}{4pt}
\centering
\caption{Ablation Study}\label{tab:ablation}
\scalebox{1.0}{
\begin{tabular}{ c  c  c }
\toprule
Settings & EV-Action & UCI \\
\midrule
Intra-view Only & 0.8146 & 0.9206 \\
Inter-view Only & 0.8036 & 0.9279 \\
Channel-aware Fusion Only & 0.8046 & 0.9095 \\
Ours without Channel-aware fusion & 0.8206 & 0.9256 \\
C$^2$AF (Ours-complete)& \textbf{0.8406} & \textbf{0.9323} \\
\bottomrule
\end{tabular}
}
\vspace{-4mm}
\end{table}
\section{Conclusions}
In this study, we have introduced a novel end-to-end Correlative Channel-Aware Fusion (C$^2$AF) network for the multi-view time series classification (MVTSC) problem. The global-local temporal encoders are developed to extract robust temporal representations for each view, and a learnable fusion mechanism is proposed to boost the multi-view label information. Extensive experimental results on three public datasets demonstrate the effectiveness of our model over the state-of-the-art methods. A detailed ablation study further validates the necessity of each component in the proposed C$^2$AF network.

\bibliographystyle{plainnat}
\bibliography{arxiv}
\end{document}